# A New Multi Criteria Decision Making Method : Approach of Logarithmic Concept (APLOCO)

Tevfik Bulut

The Ministry of Science, Industry and Technology, Ankara 6000, Turkey

*ABSTRACT*

*The primary aim of the study is to introduce APLOCO method which is developed for the solution of multi-criteria decision making problems both theoretically and practically. In this context, application subject of APLACO constitutes evaluation of investment potential of different cities in metropolitan status in Turkey.*

*The secondary purpose of the study is to identify the independent variables affecting the factories in the operating phase and to estimate the effect levels of independent variables on the dependent variable in the organized industrial zones (OIZs), whose mission is to reduce regional development disparities and to mobilize local production dynamics. For this purpose, the effect levels of independent variables on dependent variables have been determined using the multilayer perceptron (MLP) method, which has a wide use in artificial neural networks (ANNs). The effect levels derived from MLP have been then used as the weight levels of the decision criteria in APLOCO. The independent variables included in MLP are also used as the decision criteria in APLOCO. According to the results obtained from APLOCO, Istanbul city is the best alternative in term of the investment potential and other alternatives are Manisa, Denizli, Izmir, Kocaeli, Bursa, Ankara, Adana, and Antalya, respectively.*

*Although APLOCO is used to solve the ranking problem in order to show application process in the paper, it can be employed easily in the solution of classification and selection problems. On the other hand, the study also shows a rare example of the nested usage of APLOCO which is one of the methods of operation research as well as MLP used in determination of weights.*

## 1. INTRODUCTION

Decision making is the process of determining the best alternative on the basis of multiple criteria and taking into account the decision-makers' preferences. The process consists of four stages extensively; (1) determining the problem (2) introducing alternatives, (3) evaluating alternatives, and (4) identifying the best alternative [1].

Multi criteria decision analysis (MCDA) covering methods that are used to support the process of decision making deals with alternatives in both a flexible manner and an understandable structure by considering multiple criteria [2]. These methods are a branch of operations research containing mathematical tools that support the process of decision making [3]. The methods are useful for reaching an unbiased mutual idea and solving complex problems including high vagueness, conflictive purposes, different view angles, and in situations where there are many alternatives and non-objective evaluations [4,5]. At the same time, using these techniques, it can be easily to determine the best alternative, to rank alternatives, to reduce number of alternatives [6]. The most important advantage of these techniques is their ability of tackling the issues that derive from diverse clash of ideas. The techniques are being utilized progressively in the assessment of alternatives and comparative analysis [7].

Because deciding which of the MCDA methods to use is a sophisticated process, it should be considered the strengths and weaknesses of each method with regard to mathematical theory and ease of application [8].

In the paper, APLOCO developed for solving multi-criteria decision making problems is presented both theoretically and practically. In order to show the application, APLOCO has been used to evaluated the investment potential of different cities in metropolitan status. The decision criteria or variables determined for this purpose have been analyzed by MLP method which is one of the artificial neural network methods and the weight levels of these criteria have been determined and the weight levels of the criteria derived from MLP have been included in APLOCO. In the final stage, the results obtained from APLOCO and APLOCO itself are assessed in a holistic approach.

## 2. METHODOLOGY

The focus of the paper is to introduce APLOCO as a new multi-criteria decision analysis (MCDA) method and demonstrate it practically on a subject. The sub-objective of the study is to put forth the criteria to be taken into consideration for investment potential, the weight levels of these criteria and the method used for weight determination.

In this framework, input layer variables in MLP are taken as decision criteria in the APLOCO. The determined impact levels, in other words, the importance levels are used as the weights of the decision criteria in the developed the APLOCO.

### 2. 1. Data Preparation

The dataset used in the study is composed of the data for 2015 year received from the Turkish Statistical Institute (TURKSAT), the Ministry of Science, Industry and Technology (MSIT) and the Ministry of Economy (ME). The data set and resources used in the scope of the study are shown in Table 1.The SPSS 24 package programme has been used to determine the weight levels of the decision criteria and the Microsoft Office Excel 2016 program has been utilized to implement the APLOCO.

The data set of "OIZ types", "total number of parcels in production", "passing years" and "total number of factories in production" in Table 1has been compiled by me from the data of the General Directorate of Industrial Zones, which is a unit of the Ministry of Science, Industry and Technology.

**Table 1.** Data set and authorized organizations

| Name of Variable | Authorized Organizations |
|---|---|
| Total number of parcels in production [9] | MSIT |
| OIZ types [9] | |
| Passing years [9] | |
| Total number of factories in production [9] | |
| Education index [10] | TURKSTAT |
| Security index [10] | |
| Income and wealth index [10] | |
| Incentive region [11] | ME |

In the study, nine decision criteria and nine cities in metropolitan status in Turkey have been identified as alternatives, and the codes for these criteria and alternatives are presented in Table 2.

**Table 2.** Alternatives and criteria

| Alternatives | Code | Criteria | Code |
|---|---|---|---|
| Adana | A1 | Mixed OIZ | C1 |
| Ankara | A2 | Specialized OIZ | C2 |
| Antalya | A3 | Reformed OIZ | C3 |
| Bursa | A4 | Incentive zone | C4 |
| Denizli | A5 | Education index | C5 |
| Istanbul | A6 | Safety index | C6 |
| Kocaeli | A7 | Income and wealth index | C7 |
| Manisa | A8 | Total number of parcels in production | C8 |
| Izmir | A9 | Passing Years (Average) | C9 |

Criteria and alternatives for decision problem are coded and shown in Table 2. Therefore, the codes of the criteria and alternatives specified in the following sections of the study have been used.

The choice of criteria or variables has been influenced by the causality relationship between criteria or variables and the targeting of measuring the real effect. The criteria set out in this context and the reasons why these criteria are included in APLOCO and MLP are explained below.

**Criterion 1,2 and 3: Mixed, specialized and reformed OIZs as OIZ Types**

It would be appropriate to define OIZ before moving on to OIZ types.

Organized Industrial Zones (OIZs): The production areas of goods and services established to provide planned industrialization, to direct urbanization, to use resources effectively and efficiently and to establish a sustainable investment environment [12].

There are 3 different types of OIZs that operate in Turkey. These are the following:

a) Specialized OIZ: The OIZ with factories operating in a similar sector group [12].

b) Reformed OIZ: The OIZ where factories are established according to the plan to be implemented prior to April 12, 2011[13].

c) Mixed OIZ: The OIZ including factories operating in different sector group[13].

**Criterion 2: Incentive zone**

The incentive zones in Turkey are divided into six regions according to the socio-economic development index published in 2011 year [11].Incentive zones determine the proportion, amount and duration of the incentives generally provided. The rate, duration and amount of the incentives increase when it is gone from the 1st incentive zone to the 6th incentive zone [14].In here, the aim is to reduce regional development disparities with incentives provided.

Many studies in the literature show that there is a causality relationship between incentives and investment decisions and that incentives are among the optimal location selection criteria [15-20].

**Criterion 3: Education index**

Well-Being Index for Provinces, which was published by TURKSTAT for the first time in 2015, is an index study to measure, compare and monitor the life dimensions of individuals and households at the provincial level using objective and subjective criteria. The index which is represented with 41 indicators covers 11 dimensions of life in a single composite index; income and wealth, housing, health, work life, environment, education, safety, access to infrastructure services civic engagement, life satisfaction. The index is interpreted over the values between 0 and 1. The fact that the index values are close to 1 indicates that the level of well-being is better. Indicators and dimensions of the index are determined by considering the framework of OECD Better Life Index and the specific conditions in Turkey. The education index, which is the sub-index of this composite index, addresses the dimension of education that has a great importance in ensuring the knowledge and competencies necessary for individuals to participate efficiently in the society and the economy[10].

The studies show that there is a causal relationship between education and education level and entrepreneurial behavior, and that the level of education increases the likelihood of becoming an entrepreneur in high technology sectors [21-23].

**Criterion 4: Income and wealth index**

The index addresses the dimension of the income and wealth that meet the basic needs of people and provides protection against economic and personal risks, and is the sub-index of the well-being index for provinces composite index mentioned earlier [10].

It is observed that the income situation influences the entrepreneurs' capacity to establish a business and as the level of education increases in general, tendency to take risks in becoming entrepreneurs rises [24-26].

**Criterion 5: Safety index**

The safety index addresses the security dimension, which is one of the most basic needs of individuals and can be considered as a precondition for continuing other vital activities. This index is the sub-index of the well-being index mentioned earlier [10].

The studies show that the investment risk perception increases in places where the trust environment can not be established or the security measures are not taken adequately. Therefore, the high level of investment risk perception causes these places not to be invested and demographically causes the individuals in these places to migrate to other places in the country [27-30].

**Criterion 6: Total number of parcels in production**

The total number of parcels in production refers to the sum of the industrial areas that have been allocated investors in the OIZ but have not yet passed through production stage and have passed through production stage. The number of factories in production stage is a natural result of the number of parcels in production stage. Hence, it is aimed to see how effective this criterion or variable is on the number of factories passing through production stage, which is effective when this criterion or variable is included in MLP and APLOCO.

**Criterion 7: Passing years**

The "passing years" variable or criterion refers to the timeframe of the last year since the establishment of the OIZ. The reason for the inclusion of this criterion in MLP and APLOCO is due to the fact that it is desired to measure whether the elapsed time period has increased the number of factories passing through production. While this criterion has been introduced into the APLOCO, the mean values of the years from the establishment of all the OIZs to the end of 2015 year (including this year) are used.

## 2.2. The determination of the criteria weights

MLP, one of the artificial neural network methods, has been used to determine the weighting levels of the decision criteria in the APLACO. Firstly, the effect levels of independent variables affecting firms in production stage in OIZs has been determined by MLP. Subsequently, the effect levels or levels of significance of the independent variables have been used as weight levels of the decision criteria in the APLOCO. Variables used as decision criteria in determining weight levels of decision criteria and information about network architecture are shown in Table 3.In the study, all OIZs including factories in production stage until the end of 2015 year are included in the analysis.

**Table 3.** Network information in MLP

| Input Layer | Factors | 1 | Mixed |
|---|---|---|---|
| | | 2 | Specialized |
| | | 3 | Reformed |
| | | 4 | Incentive zone |
| | Covariates | 1 | Education index |
| | | 2 | Income and wealth index |
| | | 3 | Security index |
| | | 4 | Total number of parcels in production |
| | | 5 | Passing years |
| | Number of Units[a] | | 17 |
| | Rescaling Method for Covariates | | Adjusted normalized |
| Hidden Layer(s) | Number of Hidden Layers | | 1 |
| | Number of Units in Hidden Layer 1[a] | | 5 |
| | Activation Function | | Hyperbolic tangent |
| Output Layer | Dependent Variables | 1 | Total number of factories in production |
| | Number of Units | | 1 |
| | Rescaling Method for Scale Dependents | | Standardized |
| | Activation Function | | Identity |
| | Error Function | | Sum of Squares |
| a. Excluding the bias unit | | | |

Descriptive statistics in assosiciated with variables of the OIZ Type in MLP shows that out of 200 OIZs which operated by the end of 2015 year, 182 (91%), 15 (7,5%) and 3 (1,5%) were mixed, specialized and reformed types, respectively. Out of total 28,324 (100%) factories in operation phase by OIZ types in the same time period, 26,779 (91%) were in mixed ones, 1,424 (7,5%) in specialized ones and 121 (1,5%) in reformed ones. The descriptive statistics of variables apart from the variables of the OIZ type and incentive zone to be analyzed are indicated in Table 4.

**Table 4** Descriptive statistics

| Variables | N | Minimum | Maximum | Mean | Std. Deviation |
|---|---|---|---|---|---|
| Total number of parcels in production | 200 | 1,00 | 19252,00 | 241,10 | 1471,87 |
| Income and wealth index | 200 | 0,02 | 0,88 | 0,48 | 0,18 |
| Education index | 200 | 0,16 | 0,75 | 0,56 | 0,10 |
| Safety index | 200 | 0,40 | 0,79 | 0,61 | 0,10 |
| Passing years | 200 | 0,00 | 15,00 | 12,41 | 3,70 |
| Total number of factories in production | 200 | 1,00 | 7173,00 | 141,62 | 595,44 |

The normality tests have been performed in the SPSS 24 package programme to test whether the dependent variable used in the study has a normal distribution. Obtained testing results of Kolmogorov-Smirnov and Shapiro-Wilk have been indicated at Table 5.

Table 5. Normality tests

| Tests | Statistic | df | Sig. |
|---|---|---|---|
| Kolmogorov Smirnov | 0,407 | 200 | ,000 |
| Shapiro-Wilk | 0,186 | 200 | ,000 |

Because the significance level in the Shapiro-Wilk test as shown in Table 5 is below 0.05, the data doesn't show normal distribution. Similarly, that of Kolmogorov-Smirnov is not significant.

When normality test results are evaluated, it is seen that dependent variable does not show normal distribution. In these cases, an analysis method that takes into account the fact that the dependent variable is not normally distributed should be selected. Therefore, as it is in this study, MLP has been used without both doing data tranformation and excluding extreme values in cases where data are not normally distributed in the literature [31-36]. At the same time, MLP trained with back propagation which is one of the artificial intelligence algorithms have been shown to be widely used in different fields in recent years in predicting and determining correlation levels. Many studies have been conducted towards these fields of use in the industry. Some of these are following; a review: artificial neural networks as tool for control food industry process [37],parametric vs. neural network models for the estimation of production costs: a case study in the automotive industry [38], artificial neural networks workflow and its application in the petroleum industry [39], artificial neural networks applied to polymer composites: a review [40],the application of neural networks to the paper-making industry [41], suitability of multi-layer perceptron neural network model for the prediction of roll forces and motor powers in industrial hot rolling of high strength steels [42],a comprehensive review for industrial applicability of artificial neural networks [43].

Apart from these, MLP is seen to be used in many other areas. Some of these examples are; developing neural networks to investigate relationships between air quality and quality of life indicators [44], applying artificial neural network algorithms to estimate suspended sediment load (case study: Kasilian catchment, Iran) [45], importance of artificial neural network in medical

International Journal of Artificial Intelligence and Applications (IJAIA), Vol.9, No.1, January 2018

diagnosis disease like acute nephritis disease and heart disease [46], acomparative study on breast cancer prediction using RBF and MLP [47], comparative analysis of multilayer perceptron and radial basis function ANN for prediction of cycle time of structural subassembly manufacturing [48], comparison of artificial neural network, logistic regression and discriminant analysis efficiency in determining risk factors of type 2 diabetes [49], neural networks and multivariate statistical methods in traffic accident modelling [50],application of ANN to assess credit risk: a predictive model for credit card scoring [51], use of an artificial neural network to predict persistent organ failure in patients with acute pancreatitis [52], behaviour analysis of multilayer perceptrons with multiple hidden neurons and hidden layers [53] and predicting student academic performance in an engineering dynamics course: a comparison of four types of predictive mathematical models [54].

In this section, the effect levels of independent variables have been analysed by using MLP in SPSS 24 package programme. From MLP analysis, 142 cases (71,0%) have been attributed to the training sample, 58 (29,0%) to the test sample. The selection of partition records has beendone in a random fashion to develop a generalized MLP model. MLP architecture has one input layer including four factors and five covariates, one hidden layer with five units and one output layer with one dependent variable as shown in Table 3.In the model, activation function in hidden layer is hyperbolic tangent and error function is sum of squares.

In MLP model summary which is in Table 6, the relative error for the training sample is 7,9 percent. This indicates that 7,9 percent of the predictions are incorrect. The relative error for the testing sample is 1,6 percent. This implies that 4,5 percent of the predictions are incorrect. The results derived from MLP are well below the desired 10% range. This situation indicates that the results are fairly good.

**Table 6.** MLPmodel summary

| Sample | Sum of Squares Error | Relative Error (%) |
|---|---|---|
| Training | 5,601 | 7,9 |
| Testing | 8,007 | 4,5 |

From MLP analysis output, the significance levels of the independent variables (input layer variables) on the dependent variable (output layer variable), in other words, the effect levels are shown in Table 7. These effect levels has been then used as weights for decision criteria in APLOCO.

**Table 7.** Independent variable importance in MLP

| Variable | Importance Level | Rank |
|---|---|---|
| Mixed | 0,013 | 7 |
| Specialized | 0,016 | 6 |
| Reformed | 0,018 | 5 |
| Incentive zone | 0,043 | 4 |
| Education index | 0,048 | 3 |
| Income and wealth index | 0,096 | 2 |
| Safety index | 0,007 | 9 |

22

| | | |
|---|---|---|
| Total number of parcels in production | 0,750 | 1 |
| Passing years | 0,009 | 8 |

## 3. APLOCO METHOD

### 3.1. The theoretical framework of APLOCO method

APLOCO procedure is expressed in a seriesof following steps.

**Step 1: Building the decision matrix**

The generated decision matrix (X) is a cxr dimensional matrix with the criteria in the rows and the alternatives in the columns. In here, c and rrefer the number of criteria, the number of alternatives, respectively. $x_j$ indicate the decision variables and $x_i$ refers the values of the decision variables. This matrix is shown in equation (1).

$$X_{ij} = \begin{array}{c} \\ Criteria \end{array} \begin{bmatrix} C1 \\ C2 \\ \cdots \\ \cdots \\ \cdots \\ C_c \end{bmatrix} \overset{Alternatives}{\begin{bmatrix} A_1 & A_2 & \cdots & A_r \end{bmatrix}} \begin{bmatrix} x_{11} & x_{12} & \cdots & x_{1r} \\ x_{21} & x_{22} & \cdots & x_{2r} \\ \cdots & \cdots & \cdots & \cdots \\ \cdots & \cdots & \cdots & \cdots \\ \cdots & \cdots & \cdots & \cdots \\ x_{c1} & x_{c2} & \cdots & x_{cr} \end{bmatrix} = \begin{bmatrix} x_{11} & x_{12} & \cdots & x_{1r} \\ x_{21} & x_{22} & \cdots & x_{2r} \\ \cdots & \cdots & \cdots & \cdots \\ \cdots & \cdots & \cdots & \cdots \\ \cdots & \cdots & \cdots & \cdots \\ x_{c1} & x_{c2} & \cdots & x_{cr} \end{bmatrix} \quad (1)$$

**Step 2: Calculation of starting point criteria (SPC) values**

At this stage, if the criterion value is required to be maximum, the maximum value among the values of the relevant criterion in that row is determined as the maximum value. If the criterion value is desired to be minimum, the minimum value among the values of the relevant criterion in that line is determined as the minimum value. Here, if the desired condition of criterion is maximum, the criterion values in the row to which the maximum value belongs are subtracted from the maximum value. On the other hand, if the desired condition of criterion is minimum, the minimum value is subtracted from the criterion values in the row to which it belongs. In order to perform the operations mentioned, the values of $P_{ij}$ including the maximum and minimum starting point criterion values are obtained by using the equations (2), (3) and (4) respectively. In here, for $i=1,\ldots,c; j=1,\ldots,r$

$$P_{ij} = \begin{cases} \max_i x_{ij} - x_{ij} & \text{if } P_{ij} \text{ is the maximum starting point criterion.} \\ x_{ij} - \min_i x_{ij} & \text{if } P_{ij} \text{ is the minimum starting point criterion.} \end{cases} \quad (2)$$

$$P_{ij} = \begin{bmatrix} \max_i x_{ij} - x_{11} & \max_i x_{ij} - x_{12} & \cdots & \max_i x_{ij} - x_{1r} \\ \max_i x_{ij} - x_{21} & \max_i x_{ij} - x_{22} & \cdots & \max_i x_{ij} - x_{2r} \\ \cdots & \cdots & \cdots & \cdots \\ \cdots & \cdots & \cdots & \cdots \\ \cdots & \cdots & \cdots & \cdots \\ \max_i x_{ij} - x_{c1} & \max_i x_{ij} - x_{c2} & \cdots & \max_i x_{ij} - x_{cr} \end{bmatrix} = \begin{bmatrix} p_{11} & p_{12} & \cdots & p_{1r} \\ p_{21} & p_{22} & \cdots & p_{2r} \\ \cdots & \cdots & \cdots & \cdots \\ \cdots & \cdots & \cdots & \cdots \\ \cdots & \cdots & \cdots & \cdots \\ p_{c1} & p_{c2} & \cdots & p_{cr} \end{bmatrix} \quad (3)$$

$$P_{ij} = \begin{bmatrix} x_{11} - \min_i x_{ij} & x_{12} - \min_i x_{ij} & \cdots & x_{1r} - \min_i x_{ij} \\ x_{21} - \min_i x_{ij} & x_{22} - \min_i x_{ij} & \cdots & x_{2r} - \min_i x_{ij} \\ \cdots & \cdots & \cdots & \cdots \\ \cdots & \cdots & \cdots & \cdots \\ \cdots & \cdots & \cdots & \cdots \\ x_{c1} - \min_i x_{ij} & x_{c2} - \min_i x_{ij} & \cdots & x_{cr} - \min_i x_{ij} \end{bmatrix} = \begin{bmatrix} p_{11} & p_{12} & \cdots & p_{1r} \\ p_{21} & p_{22} & \cdots & p_{2r} \\ \cdots & \cdots & \cdots & \cdots \\ \cdots & \cdots & \cdots & \cdots \\ \cdots & \cdots & \cdots & \cdots \\ p_{c1} & p_{c2} & \cdots & p_{cr} \end{bmatrix} \quad (4)$$

**Step 3: Forming the logarithmic conversion (LC) matrix**

In this stage, the +2 integer value is added to each of the $P_{ij}$ values $(p_{11}, p_{12}, p_{13}, \ldots p_{1r})$ in the rows. After this, the LC values are obtained by calculating the natural logarithm of the multiplicative inverse of the obtained results. This operation is performed with the help of equation (5) and the normalization process is completed and then the logarithmic transformation matrix in equation (6) is obtained. Here, ln is undefined when it receive 0 and negatives values and is equal to 0 when ln takes 1. For these reasons, +2, which is greater than 1 in equation (4), is taken as an integer value. Another reason for adding the number 2 as an integer value to the natural logarithm value is to avoid extreme values and negative values and to ensure that the values are positive.

$$\ln(x) = \log_e(x) \text{ and } L_{ij} = \frac{1}{\ln(p_{ij} + 2)} \text{ for } (i = 1 \ldots c \text{ and } j = 1 \ldots r) \quad (5)$$

$$L_{ij} = \begin{bmatrix} \frac{1}{\ln(p_{11} + 2)} & \frac{1}{\ln(p_{12} + 2)} & \cdots & \frac{1}{\ln(p_{1r} + 2)} \\ \frac{1}{\ln(p_{21} + 2)} & \frac{1}{\ln(p_{22} + 2)} & \cdots & \frac{1}{\ln(p_{2r} + 2)} \\ \cdots & \cdots & \cdots & \cdots \\ \cdots & \cdots & \cdots & \cdots \\ \cdots & \cdots & \cdots & \cdots \\ \frac{1}{\ln(p_{c1} + 2)} & \frac{1}{\ln(p_{c2} + 2)} & \cdots & \frac{1}{\ln(p_{cr} + 2)} \end{bmatrix} = \begin{bmatrix} l_{11} & l_{12} & \cdots & l_{1r} \\ l_{21} & l_{22} & \cdots & l_{2r} \\ \cdots & \cdots & \cdots & \cdots \\ \cdots & \cdots & \cdots & \cdots \\ \cdots & \cdots & \cdots & \cdots \\ l_{c1} & l_{c2} & \cdots & l_{cr} \end{bmatrix} \quad (6)$$

## Step 4: Determining the weights of criteria and calculating the weighted logarithmic conversion (WLC) matrix

$W = [w_1, w_2, w_3, ...., w_n]$ is a weight coefficient. In here, $w_1 \in \Re$ and $\sum_{i=1}^{n} w_i = 1$. The WLC matrix in equation (7) is obtained by multiplying the logarithmic transformed $l_{ij}$ values by the weight levels ($w_i$) of the criteria determined by any method.

$$T_{ij} = \begin{bmatrix} l_{11}w_1 & l_{12}w_1 & ... & l_{1r}w_1 \\ l_{21}w_2 & l_{22}w_2 & ... & l_{2r}w_2 \\ ... & ... & ... & ... \\ ... & ... & ... & ... \\ ... & ... & ... & ... \\ l_{c1}w_n & l_{c2}w_n & ... & l_{cr}w_n \end{bmatrix} = \begin{bmatrix} t_{11} & t_{12} & ... & t_{1r} \\ t_{21} & t_{22} & ... & t_{2r} \\ ... & ... & ... & ... \\ ... & ... & ... & ... \\ ... & ... & ... & ... \\ t_{c1} & t_{c2} & ... & t_{cr} \end{bmatrix} \quad (7)$$

## Step 5: Determination of the best alternative

The maximum values of the criteria in each row are determined as the optimal solution values ($\beta_j$) and the scores ($\beta_{sj}$) are obtained after the sum is taken. This operation is done by means of equations (8) and (9), respectively. The final scores ($\theta_i$) of each alternative are calculated by proportioning the total scores ($\alpha_{si}$) of the criterion values of the alternatives to the optimal solution values ($\beta_{sj}$) summed. This operation is done by equations (10) and (11), respectively. The scores obtained from Equation (11) are between 0 and 1, allowing the scores to be evaluated in this range. Then, $\theta_i$ values are then sorted from large to small, and the first-order alternative is considered the most appropriate alternative.

Similarly, the difference between the $\alpha_{si}$ scores of the alternatives and the $\beta_{sj}$ score expressing the sum of the optimal solution values shows how far $\alpha_{si}$ scores are from the optimal solution values. Among the distances to be shown on the graph, the alternative with the closest scorer to the score $\beta_{sj}$ is determined as the best alternative.

$\beta_j = \{\max_i t_{ij}\}$ and $\beta_j = \{t_1, t_2, t_3, ......, t_{1n}\}$ is the maximum value of the criterion on each row. (8)

$\beta_{sj} = \sum_{j=1}^{n} (l_1, l_2, l_3 .... l_n)$ refers sum of the maximum values of the criteria. (9)

$a_{si} = \sum_{j=1}^{n} (t_1, t_2, t_3 .... t_n)$ indicates the sum of values of the criteria for each alternative in each column. (10)

$0 \leq \theta_i \leq 1$ and $j = 1, 2, ...., r$. $\theta_i = \dfrac{a_{si}}{\beta_{sj}}$ refers the final scores of each alternative. (11)

## 3.2. Application Results

In this section, nine different provinces in metropolitan status have been evaluated by APLOCO in terms of the investment potential based on the variables affecting factories in operation phase in OIZs. The implementation steps of the method are completed in five steps.

**Step 1: Building decision matrix**

The first step of APLOCO begins with the creation of the decision matrix. In this context, a decision matrix consisting of nine previously determined decision criteria and alternatives has been established. This generated matrix is indicated in Table 8.

Table 8. Decision matrix

|    | A1    | A2       | A3    | A4     | A5    | A6       | A7    | A8    | A9     |
|----|-------|----------|-------|--------|-------|----------|-------|-------|--------|
| C1 | 2,00  | 7,00     | 1,00  | 9,00   | 2,00  | 7,00     | 6,00  | 4,00  | 9,00   |
| C2 | 0,00  | 0,00     | 0,00  | 2,00   | 1,00  | 1,00     | 5,00  | 0,00  | 0,00   |
| C3 | 0,00  | 0,00     | 0,00  | 1,00   | 0,00  | 0,00     | 0,00  | 0,00  | 0,00   |
| C4 | 2,00  | 1,00     | 1,00  | 1,00   | 2,00  | 1,00     | 1,00  | 3,00  | 1,00   |
| C5 | 0,45  | 0,55     | 0,64  | 0,60   | 0,65  | 0,52     | 0,58  | 0,53  | 0,60   |
| C6 | 0,42  | 0,47     | 0,41  | 0,62   | 0,51  | 0,47     | 0,48  | 0,62  | 0,52   |
| C7 | 0,35  | 0,80     | 0,58  | 0,54   | 0,54  | 0,88     | 0,63  | 0,39  | 0,66   |
| C8 | 383,00| 11246,00 | 216,00| 1335,00| 142,00| 22665,00 | 796,00| 302,00| 1249,00|
| C9 | 14,00 | 13,00    | 14,00 | 10,00  | 15,00 | 14,00    | 13,00 | 12,00 | 13,00  |

**Step 2: Calculation of starting point criteria (SPC) values**

When the values to which the decision criteria are taken are evaluated in terms of the best starting point values, it is specified as the maximum the conditions of the following criteria; mixed and specialized OIZs, incentive zone, education index, income and wealth index, security index, total number of parcels in production and passing years. The reason for this is that these criteria with the high condition provides the greatest benefit. Here, the maximum values of the criteria in the rows are determined because the desired criterion conditions are maximum (Max).Subsequently, all the criterion values in the desired rows with the maximum condition are subtracted from these maximum values and the maximum starting point criteria values are obtained.

The reformed OIZs refer to OIZs that do not meet OIZ requirements at first. However, if these OIZs provide OIZ requirements, they are candidates to become OIZ. If the reformed OIZs provide reformed requirements for the period specified in the OIZ Law, they can only be a OIZ in a real sense [55].For these reasons, the criterion condition of the reformed OIZ is taken as minimum (Min).Since the desired criterion condition is the minimum, the minimum values of the criteria are first found in the rows. Subsequently, these minimum values are subtracted from all criteria values in the desired rows with the minimum condition and the minimum starting point criteria values are obtained.

The matrix in Table 9 are formed by obtaining the $P_{ij}$ values of as a result of the above-mentioned operations

**Table 9.** The SPC values

|    | Min and Max | A1 | A2 | A3 | A4 | A5 | A6 | A7 | A8 | A9 |
|----|------|------|------|------|------|------|------|------|------|------|
| C1 | Max | 7,00 | 2,00 | 8,00 | 0,00 | 7,00 | 2,00 | 3,00 | 5,00 | 0,00 |
| C2 | Max | 5,00 | 5,00 | 5,00 | 3,00 | 4,00 | 4,00 | 0,00 | 5,00 | 5,00 |
| C3 | Min | 0,00 | 0,00 | 0,00 | 1,00 | 0,00 | 0,00 | 0,00 | 0,00 | 0,00 |
| C4 | Max | 1,00 | 2,00 | 2,00 | 2,00 | 1,00 | 2,00 | 2,00 | 0,00 | 2,00 |
| C5 | Max | 0,20 | 0,10 | 0,01 | 0,05 | 0,00 | 0,14 | 0,07 | 0,12 | 0,05 |
| C6 | Max | 0,20 | 0,16 | 0,21 | 0,00 | 0,11 | 0,15 | 0,15 | 0,00 | 0,10 |
| C7 | Max | 0,53 | 0,08 | 0,30 | 0,34 | 0,34 | 0,00 | 0,25 | 0,48 | 0,22 |
| C8 | Max | 22282,00 | 11419,00 | 22449,00 | 21330,00 | 22523,00 | 0,00 | 21869,00 | 22363,00 | 21416,00 |
| C9 | Max | 1,00 | 2,00 | 1,00 | 5,00 | 0,00 | 1,00 | 2,00 | 3,00 | 2,00 |

**Step 3: Forming the logarithmic conversion (LC) matrix**

In this phase, the +2 integer value is added to the criterion values in each row. Next, the LC matrix in Table 10 is obtained by calculating the natural logarithm of the multiplicative inverse of the obtained results.

**Table 10.** The LCmatrix

|    | A1 | A2 | A3 | A4 | A5 | A6 | A7 | A8 | A9 |
|----|------|------|------|------|------|------|------|------|------|
| C1 | 0,46 | 0,72 | 0,43 | 1,44 | 0,46 | 0,72 | 0,62 | 0,51 | 1,44 |
| C2 | 0,51 | 0,51 | 0,51 | 0,62 | 0,56 | 0,56 | 1,44 | 0,51 | 0,51 |
| C3 | 1,44 | 1,44 | 1,44 | 0,91 | 1,44 | 1,44 | 1,44 | 1,44 | 1,44 |
| C4 | 0,91 | 0,72 | 0,72 | 0,72 | 0,91 | 0,72 | 0,72 | 1,44 | 0,72 |
| C5 | 1,26 | 1,35 | 1,43 | 1,40 | 1,44 | 1,32 | 1,37 | 1,33 | 1,39 |
| C6 | 1,27 | 1,30 | 1,26 | 1,44 | 1,34 | 1,31 | 1,31 | 1,44 | 1,35 |
| C7 | 1,08 | 1,36 | 1,20 | 1,18 | 1,18 | 1,44 | 1,23 | 1,10 | 1,25 |
| C8 | 0,10 | 0,11 | 0,10 | 0,10 | 0,10 | 1,44 | 0,10 | 0,10 | 0,10 |
| C9 | 0,91 | 0,72 | 0,91 | 0,51 | 1,44 | 0,91 | 0,72 | 0,62 | 0,72 |

**Step 4: Determining the weights of criteria and calculating the weighted logarithmic conversion (WLC) matrix**

In this step, the WLC matrix in Table 11 is obtained by multiplying the criterion values in Table 10 by the criterial weights determined in MLP method.

**Table 11.** The WLCmatrix

|    | Weight | A1 | A2 | A3 | A4 | A5 | A6 | A7 | A8 | A9 |
|----|------|------|------|------|------|------|------|------|------|------|
| C1 | 0,013 | 0,006 | 0,010 | 0,006 | 0,019 | 0,006 | 0,010 | 0,008 | 0,007 | 0,019 |
| C2 | 0,016 | 0,008 | 0,008 | 0,008 | 0,010 | 0,009 | 0,009 | 0,024 | 0,008 | 0,008 |
| C3 | 0,018 | 0,026 | 0,026 | 0,026 | 0,016 | 0,026 | 0,026 | 0,026 | 0,026 | 0,026 |

| | | | | | | | | | |
|---|---|---|---|---|---|---|---|---|---|
| C4 | 0,043 | 0,039 | 0,031 | 0,031 | 0,031 | 0,039 | 0,031 | 0,031 | 0,062 | 0,031 |
| C5 | 0,048 | 0,060 | 0,064 | 0,068 | 0,066 | 0,069 | 0,063 | 0,065 | 0,063 | 0,066 |
| C6 | 0,096 | 0,122 | 0,125 | 0,121 | 0,139 | 0,129 | 0,126 | 0,126 | 0,139 | 0,130 |
| C7 | 0,007 | 0,007 | 0,009 | 0,008 | 0,008 | 0,008 | 0,009 | 0,008 | 0,007 | 0,008 |
| C8 | 0,750 | 0,075 | 0,080 | 0,075 | 0,075 | 0,075 | 1,082 | 0,075 | 0,075 | 0,075 |
| C9 | 0,009 | 0,008 | 0,006 | 0,008 | 0,004 | 0,013 | 0,008 | 0,006 | 0,005 | 0,006 |

**Step 5: Determination of the best alternative**

After the maximum values of the criteria in each row are determined as the optimal solution values, the maximum values of the criteria are summed and the $\beta_{sj}$ score is obtained as shown in Table 12.

**Table 12.** The values of the optimal solution criteria

| Criteria | C1 | C2 | C3 | C4 | C5 | C6 | C7 | C8 | C9 | $\beta_{sj}$ Score |
|---|---|---|---|---|---|---|---|---|---|---|
| Values | 0,019 | 0,024 | 0,026 | 0,062 | 0,069 | 0,139 | 0,009 | 1,082 | 0,013 | 1,443 |

Then, the $\alpha_{si}$ scores are calculated by summing the criterion values of alternatives in each column. The $\alpha_{si}$ scores of the alternatives are proportioned to the $\beta_{sj}$ score and the best alternative are determined according to the $\theta_i$ scores obtained as shown in Table 13. Thanks to this operation, the obtained $\theta_i$ scores are between 0 and 1 and the scores are evaluated within this range.

**Table 13.** The scores of the alternatives

| Code | Alternatives | $\alpha_{si}$ Scores | $\beta_{sj}$ Score | $\theta_i$ Scores | Rank of $\theta_i$ Scores |
|---|---|---|---|---|---|
| A1 | Adana | 0,352 | 1,443 | 0,244 | 8 |
| A2 | Ankara | 0,360 | 1,443 | 0,249 | 7 |
| A3 | Antalya | 0,351 | 1,443 | 0,244 | 9 |
| A4 | Bursa | 0,370 | 1,443 | 0,256 | 6 |
| A5 | Denizli | 0,373 | 1,443 | 0,258 | 3 |
| A6 | Istanbul | 1,364 | 1,443 | 0,945 | 1 |
| A7 | Kocaeli | 0,370 | 1,443 | 0,256 | 5 |
| A8 | Manisa | 0,393 | 1,443 | 0,272 | 2 |
| A9 | Izmir | 0,371 | 1,443 | 0,257 | 4 |

On the other hand, the difference between the $\beta_{sj}$ score and the $\alpha_{si}$ score of alternatives shows how far the $\alpha_{si}$ scores are away from the $\beta_{sj}$ score, the sum of the optimal solution values. Among the obtained distances, the alternative with the closest one to the $\beta_{sj}$ score is the best alternative at the same time. In this context, the results in Table 13 can also be summarized via Figure 1. According to Figure 1 showing the distance of the $\alpha_{si}$ scores of the alternatives to the $\beta_{sj}$ score and Table 13 showing the scores of $\alpha_{si}$, $\beta_{sj}$ and $\theta_i$, the alternative A6 (Istanbul) which is closest to the $\beta_{sj}$ score among the alternatives is the best alternative from the investment potential point of view and other alternatives are Manisa (A8), Denizli (A5), Izmir (A9), Kocaeli (A7), Bursa (A4), Ankara (A2), Adana (A1) and Antalya (A3), respectively.

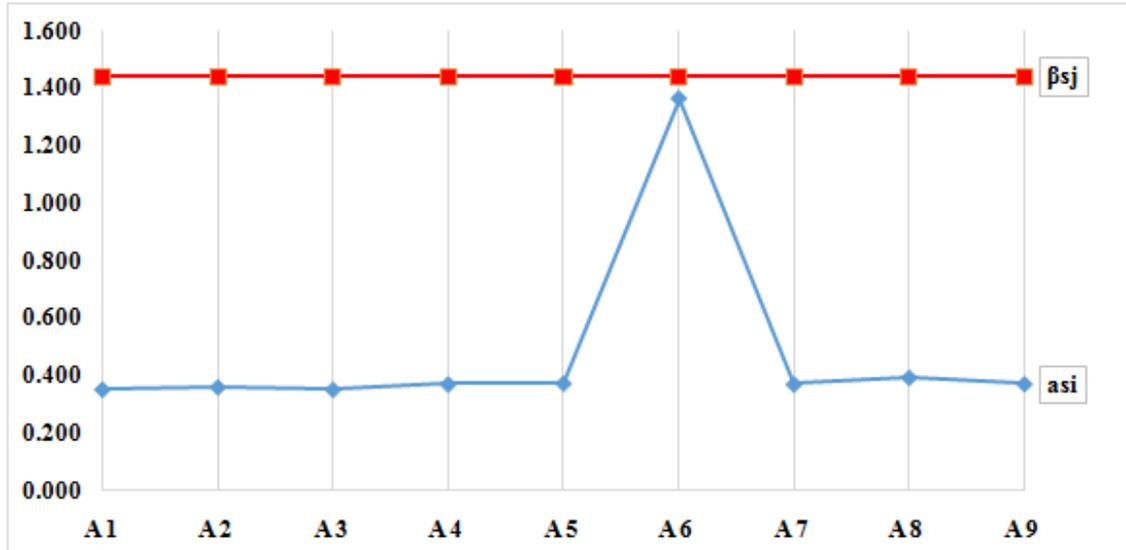

**Figure 1.** Distances of the $\alpha_{si}$ scores from the $\beta_{sj}$ score

## 4. CONCLUSION

In the research, the variables affecting the factories at the operational phase in the OIZs and the importance levels of these variables are determined by MLP which is one of the artificial neural network methods. The variables whose importance level is determined are used as decision criteria and the importance levels of these variables are taken as weight levels of decision criteria. Nine different metropolitan cities have been evaluated in terms of investment potential of provinces using APLOCO taking into account these decision criteria and the weight levels of these decision criteria that have been determined. According to the results obtained from APLOCO, the Istanbul province (A6) alternative has been assigned to the 1st place among 9 provinces in terms of investment potential. The province of Istanbul that has been assigned to the 1st place has been followed by Manisa (A8), Denizli (A5), Izmir (A9), Kocaeli (A7), Bursa (A4), Ankara (A2), Adana (A1) and Antalya (A3), respectively. It has been determined that Istanbul, which is ranked in the first place in the investment potential ranking of the provinces, is more advantageous than the others in terms of the total number of parcels in production criterion.

Especially in cases where data are not normally distributed, MLP has been widely used in recent years and in some studies it is seen that MLP has better results than other statistical methods. In this study, the results obtained from MLP have not only a weighting of the decision criteria but also a great importance in term of determining the effect levels of the independent variables on the dependent variables. Today, the number of OIZs reaching 301 has been a vital role in the development of the planned industry in Turkey [10].The estimation of the variables affecting the factories in the operating stage in the OIZs and the determination of their impact levels have of great importance in relation to establishment of a sustainable investment environment, policy development through more robust outputs and productivity increase.

Although MLP is used as the weight determination method in this study, other methods can also be used easily as an alternative to this method and weight levels can be employed as weight levels of decision criteria in APLOCO. This situation depends entirely on the decision makers' preferences to investigate or implement and structure of data set.

APLOCO which is developed is to identify the best alternatives from a given set of alternatives, can easily also be implemented in the solution of the problems including choosing, ranking and sorting in terms of both in terms of easiness of usage and requiring fewer math operations.

**Limitations of the research**

Because this study shows the character of case study, findings of the study can not be generalized to other cities and populations with different characteristics. At the same time, it is aimed to present the developed model together with the combination of MLP, which is used as the weight determination method, and it is not included the comparative results of APLOCO method with other operational research methods in this paper.

**Suggestions for future studies**

Future studies should be compared APLOCO method with other operational research methods and shown advantages and disadvantages relative to each other. In addition, the expansion of the target population, including other cities in the future study, will be important in terms of the generalization of research findings and the view of the overall picture as a whole.

*Data accessibility.*

1. Yearly integrated data for 2015:
http://www.turkstat.gov.tr/VeriBilgi.do?alt_id=1106.

2. Incentive Zones:
https://www.economy.gov.tr/portal/content/conn/UCM/path/Contribution%20Folders/web_en/Investment%20Incentives/Investment%20Incentive%20System_1492085002899.pdf?lve.

*Competing interests.* The author declares no competing financial interests.

*Funding.* The author reports that no funds are used in the research.

*Acknowledgements.* The author would like to thank Serkan Duman for providing access to in house database.

*Research Ethics.* The author says that it is not required to complete an ethical assessment prior to conducting the research because the research does not include ethical elements.

*Permission to carry out fieldwork.* The author declares that it is not required to obtain the appropriate permissions and licences to conduct the research.

```
```

## AUTHOR

Mr. Tevfik Bulut

Industry and Technology Specialist
at Republic of Turkey Ministry of Science, Industry and Technology

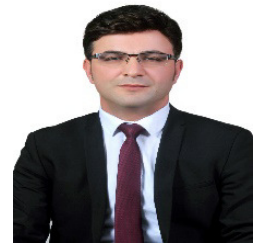